\documentclass[10pt,conference]{IEEEtran}

\usepackage[utf8]{inputenc}
\usepackage[dvipsnames]{xcolor}
\usepackage{graphicx}
\usepackage{multirow}
\usepackage{makecell}
\usepackage{amsmath}
\usepackage[caption=false,font=footnotesize]{subfig}

\DeclareMathOperator{\sign}{sign}
\DeclareMathOperator{\clip}{clip}

\title{Adversarial Robustness Assessment of NeuroEvolution Approaches}
\author{
\IEEEauthorblockN{Inês Valentim, Nuno Lourenço, Nuno Antunes}
\IEEEauthorblockA{\textit{University of Coimbra, CISUC, DEI}\\
Coimbra, Portugal\\
valentim@dei.uc.pt, naml@dei.uc.pt, nmsa@dei.uc.pt}
}

\begin{document}

\maketitle

\begin{abstract}

NeuroEvolution automates the generation of Artificial Neural Networks through the application of techniques from Evolutionary Computation.
The main goal of these approaches is to build models that maximize predictive performance, sometimes with an additional objective of minimizing computational complexity.
Although the evolved models achieve competitive results performance-wise, their robustness to adversarial examples, which becomes a concern in security-critical scenarios, has received limited attention.
In this paper, we evaluate the adversarial robustness of models found by two prominent NeuroEvolution approaches on the CIFAR-10 image classification task: DENSER and NSGA-Net.
Since the models are publicly available, we consider white-box untargeted attacks, where the perturbations are bounded by either the $L_{2}$ or the $L_{\infty}$-norm.
Similarly to manually-designed networks, our results show that when the evolved models are attacked with iterative methods, their accuracy usually drops to, or close to, zero under both distance metrics.
The DENSER model is an exception to this trend, showing some resistance under the $L_{2}$ threat model, where its accuracy only drops from 93.70\% to 18.10\% even with iterative attacks.
Additionally, we analyzed the impact of pre-processing applied to the data before the first layer of the network. Our observations suggest that some of these techniques can exacerbate the perturbations added to the original inputs, potentially harming robustness. Thus, this choice should not be neglected when automatically designing networks for applications where adversarial attacks are prone to occur.

\end{abstract}


\section{Introduction}
\label{sec:introduction}

The design of Artificial Neural Networks (ANNs) is a time-consuming trial-and-error process that requires domain expertise. The design choices affect one another and this interdependence must be considered when dealing with increasingly complex models which often reach thousands or millions of parameters. To alleviate these issues, NeuroEvolution (NE) uses evolutionary approaches to automate the search for better topologies and parametrization of ANNs.

Models designed through NE achieve competitive results in comparison with manually-designed ANNs, sometimes even surpassing their performance~\cite{assuncao2018denser,liu2018hierarchical,10.1145/3321707.3321729,Real_Aggarwal_Huang_Le_2019,10.1145/3071178.3071229}.
Thus, it is reasonable to consider that these models can be applied in real-world scenarios.
However, their adoption in such cases means that concerns other than predictive performance must be addressed.
In the particular case of security-critical systems, such as autonomous vehicles or malware detectors, one of such concerns is the models' vulnerability to adversarial examples~\cite{DBLP:journals/corr/GoodfellowSS14,DBLP:journals/corr/SzegedyZSBEGF13}.
These malicious examples are usually crafted by adding small perturbations to the original inputs causing the attacked model to produce an incorrect output~\cite{DBLP:conf/sp/Carlini017,DBLP:conf/cvpr/DongFYPSXZ20}. Restrictions are often imposed on the maximum perturbation that can be added so as to try to keep the adversarial example and the original input indistinguishable~\cite{DBLP:conf/sp/Carlini017,DBLP:conf/cvpr/DongFYPSXZ20}.
In the image domain, the perturbations are often bounded by some $L_{p}$-norm~\cite{DBLP:conf/sp/Carlini017,carlini2019evaluating}.
ANNs designed by humans have been widely studied and are known to be vulnerable to these attacks, but it is still unclear the extent to which ANNs obtained through the application of NE suffer from this vulnerability.

Our goal is to empirically evaluate the resistance of ANNs designed by NE to adversarial examples, with a focus on image classification tasks and Convolutional Neural Networks (CNNs).
We consider threat models where the attacker's goal is simply to cause a misclassification of a given instance, i.e., we perform untargeted attacks, and has full access to the models, i.e., we perform white-box attacks.
Similar to previous work in adversarial machine learning, we consider $L_{2}$ and $L_{\infty}$-bounded perturbations~\cite{croce2020robustbench}, and white-box attacks that maximize the loss function while constraining the perturbations to a pre-defined budget under each distance metric.

Our analysis is performed by attacking pre-trained models for the CIFAR-10 dataset~\cite{krizhevsky2009learning}, made publicly available by the authors of the corresponding NE approaches.
We consider DENSER~\cite{assuncao2018denser} and two variants of NSGA-Net models~\cite{10.1145/3321707.3321729}, all of which achieving an accuracy above 93\% on clean images.
These models, and their training, do not incorporate any defense mechanism against adversarial examples.
Thus, any robustness that is shown can be likely attributed to architectural aspects of the ANNs.
Moreover, the NE approaches only seek to maximize predictive performance, while directly or indirectly minimizing computational complexity.

Our results show that these evolved models are susceptible to adversarial attacks, similar to hand-designed ANNs. Using iterative methods, like the basic iterative method (BIM)~\cite{DBLP:journals/corr/KurakinGB16} and the projected gradient descent (PGD) attack~\cite{madry2018towards}, their accuracy drops dramatically, to values below 0.25\% under both distance metrics.
DENSER deviates from this behavior and shows some resistance to $L_{2}$ perturbations, keeping its accuracy at 21.76\% under a BIM attack with 100 iterations. When reducing the iterations to 50 but incorporating 10 random restarts, the accuracy falls further to 18.10\%.
On the other hand, the models that use the architectures found by NSGA-Net, especially when applied to the NASNet search space, are more robust to the single-step attacks of our experiments.
We could also identify distinct patterns in the misclassifications produced by the models, even under the same attack and when the accuracy of all of them is zero. For instance, some misclassify the adversarial examples of a class into a small subset of classes, while the misclassifications of other models are much more spread out.

Furthermore, we warn about the choice of data pre-processing procedures and the effect of such transformations on the perturbations added to the clean inputs. Certain techniques can exacerbate these perturbations before the data reaches the first layer of the network, which in a way makes it easier for the attacker to succeed in generating an adversarial example under a certain perturbation budget. For this reason, it might be worth including this component in the search process of NE approaches. Moreover, all training conditions, including any model-specific pre-processing step, should be clearly specified when evaluating adversarial robustness, especially when comparing models from different sources.

The remainder of this paper is structured as follows. Section~\ref{sec:background-related-work} provides an overview of NE approaches and adversarial machine learning, and presents relevant work on the intersection of the two fields. Section~\ref{sec:experimental-setup} describes our experimental setup in terms of datasets, models, adversarial attacks, and evaluation metrics. Section~\ref{sec:results-discussion} presents the results of the experimental campaign and discusses our main findings. Section~\ref{sec:conclusion} concludes the paper.

\section{Background and Related Work}
\label{sec:background-related-work}

In this section, we overview key concepts and methods from the fields of NeuroEvolution (NE) and adversarial machine learning, after which we present relevant work on the intersection of these two areas of research. We mainly focus on image classification and Convolutional Neural Networks (CNNs).

\subsection{NeuroEvolution}
\label{ssec:neuroevolution}

Neural Architecture Search (NAS) is the field which deals with designing the optimal architecture of an ANN in an automated way. Different search strategies have been proposed, including Reinforcement Learning~\cite{pmlr-v80-pham18a,8578355,DBLP:journals/corr/ZophL16}, Evolutionary Computation~\cite{assuncao2018denser,liu2018hierarchical,10.1145/3321707.3321729,Real_Aggarwal_Huang_Le_2019,10.1145/3071178.3071229,8237416}, as well as gradient-based methods~\cite{liu2018darts}.

NeuroEvolution (NE) approaches, which are the main focus of this work, refer to those which apply techniques from Evolutionary Computation. Especially since 2017, remarkable results have been achieved by several NE approaches that automate the search for CNNs~\cite{assuncao2018denser,liu2018hierarchical,10.1145/3321707.3321729,Real_Aggarwal_Huang_Le_2019,10.1145/3071178.3071229,8237416}. Two of such proposals, DENSER~\cite{assuncao2018denser} and NSGA-Net~\cite{10.1145/3321707.3321729}, are described in more detail in what follows.

In DENSER~\cite{assuncao2018denser}, each candidate solution is represented at two levels, the genetic algorithm (GA) level and the dynamic structured grammatical evolution (DSGE)~\cite{DBLP:books/sp/18/0001APCM18} level. The GA level encodes the macro structure of the ANN (layers and order in which they are connected) and any additional hyperparameters (learning strategy and data augmentation procedures, for instance) as a sequence of evolutionary units. Each evolutionary unit stores a non-terminal symbol later used at the DSGE level, as well as the minimum and maximum number of times the unit can be used. At the DSGE level, the parameters of the evolutionary units (e.g., layer type, number of filters, filter size, activation functions) are encoded through a context-free grammar. DENSER uses crossover and mutation operators at both levels.

NSGA-Net~\cite{10.1145/3321707.3321729} is explicitly designed to solve a multi-objective optimization problem. It uses the NSGA-II algorithm~\cite{DBLP:conf/ppsn/DebAPM00} to maintain a trade-off frontier between candidate solutions that maximize classification performance but minimize computational complexity (defined by the number of floating-point operations in the forward pass).
Each candidate solution is represented by a bit string that encodes computational blocks (referred to as \textit{phases}). In turn, each computational block can be regarded as a directed acyclic graph of several \textit{nodes}, with each node representing a single operation (e.g., convolution, pooling, batch-normalization) or a sequence of operations. This representation follows the method from~\cite{8237416} with a minor modification.
Similar to DENSER, NSGA-Net also makes use of crossover and mutation operators. Additionally, at the end of the evolutionary process, ANNs are designed by sampling phases from a Bayesian Network which models the relationship between these computational blocks as seen in the architectures found during the search procedure.

In practice, NSGA-Net constrains the search space based on prior knowledge of successful architectures. Namely, the number of phases and the maximum number of nodes on each phase are set to 3 and 6, respectively, and the changes in spatial resolution between phases are also fixed. Moreover, each node encompasses the same sequence of operations.
When it comes to DENSER, the main restriction is at the GA level in how the different layers can be connected (i.e., layers are connected sequentially without skip-connections). In the original work, the search space includes CNNs of up to 40 hidden-layers (30 convolution or pooling layers, at most, followed by a maximum of 10 fully-connected layers).

\subsection{Adversarial Machine Learning}
\label{ssec:adv-machine-learning}

Adversarial examples are maliciously crafted so as to make an attacked model produce incorrect outputs~\cite{DBLP:conf/cvpr/DongFYPSXZ20,DBLP:journals/corr/abs-1903-10484}. Throughout the years, several methods have been proposed in the literature to craft such attacks, under different threat models. These threat models can be defined based on the goals, knowledge, and capabilities of the adversary~\cite{carlini2019evaluating}.

A distinction can be made based on the adversary's knowledge about the model: architecture and parameters, training algorithm and training data, randomness at test-time, and allowed level of query access~\cite{pmlr-v80-athalye18a}. In this work, we focus on \textit{white-box attacks}, where the adversary has full access to the model.

A further distinction can be made between untargeted and targeted attacks. In the particular case of image classification tasks, the goal of \textit{untargeted attacks} is simply to make the model predict a class different from the true label of a given instance, while the goal of \textit{targeted attacks} is to make the model produce a misclassification into some desired class~\cite{DBLP:conf/sp/Carlini017}.
Formally, $x$ is an input with correct label $y$ and $C$ is the classifier under attack. In the untargeted setting, an adversarial example $x^{adv}$ is such that $C(x^{adv}) \neq y$. On the other hand, a targeted attack would aim at crafting $x^{adv}$ given a target $t \neq y$ such that $C(x^{adv}) = t$.

An adversarial example is usually obtained by adding some perturbation $\delta$ to a benign image~\cite{DBLP:journals/corr/abs-1903-10484}. Constraints are usually imposed on the capabilities of the adversary in terms of the maximum perturbation that can be added, so that $x^{adv}$ remains close to the original input and its true label remains unchanged~\cite{carlini2019evaluating}.
In the image domain, a common approach is to use $L_p$-norms for those bounds such that $\left\|x - x^{adv}\right\|_{p} \leq \epsilon$, where $\epsilon$ is the perturbation budget and usually $p \in \left\{0, 1, 2, \infty \right\}$~\cite{carlini2019evaluating,Dong2018MIM}.

Optimization-based methods are an important category of white-box attacks. Some of such methods try to minimize the perturbation, while others try to maximize some loss function (typically the cross-entropy loss)~\cite{pmlr-v80-athalye18a,Dong2018MIM}.
We present some of the latter methods assuming an untargeted setting.

The fast gradient sign method (FGSM)~\cite{DBLP:journals/corr/GoodfellowSS14} is a one-step gradient-based attack optimized for the $L_{\infty}$-norm which generates an adversarial example as:

$$ x^{adv} = x + \epsilon \cdot \sign \left(\nabla_{x} L(x, y)\right)$$

where $\nabla_{x} L(x, y)$ is the gradient of the cross-entropy loss with respect to the input image. When this method is optimized for the $L_{2}$-norm, we obtain the fast gradient method (FGM) which generates an adversarial example as:

$$ x^{adv} = x + \epsilon \cdot \frac{\nabla_{x} L(x, y)}{\left\|\nabla_{x} L(x, y)\right\|_{2}}$$

In both cases, the perturbed image should be clipped in order to maintain a valid data range.

We present the methods that follow for $L_{\infty}$ adversaries, but these definitions can be easily adapted to the case where $L_{2}$ bounds are imposed, similar to what is done with FGM.

A straightforward extension of the FGSM attack is to take multiple small steps (with step size $\alpha$) and clip the result by $\epsilon$ at each iteration. The clipping operation also takes into account the valid range of data values. This results in the basic iterative method (BIM)~\cite{DBLP:journals/corr/KurakinGB16} which can be defined as:

$$x^{adv}_{0} = x,$$
$$x^{adv}_{i+1} = \clip_{\epsilon}\left\{x^{adv}_{i} + \alpha \cdot \sign \left(\nabla_{x} L(x^{adv}_{i}, y)\right)\right\}$$

The projected gradient descent (PGD) method~\cite{madry2018towards} is another iterative attack which only differs from BIM on how $x^{adv}_{0}$ is set. In the case of PGD, instead of starting from the original input, a random perturbation bounded by $\epsilon$ is generated and added to $x$.

In an attempt to stabilize the update directions and escape from local optima, the momentum iterative fast gradient sign method (MI-FGSM)~\cite{Dong2018MIM} incorporates momentum~\cite{pmlr-v28-sutskever13} into the BIM method.
The Auto-PGD (APGD) method~\cite{pmlr-v119-croce20b} is a variation of the PGD attack which adjusts the step size in an automated way. The authors of APGD also proposed an alternative to the cross-entropy loss called difference of logits ratio (DLR) loss. In addition to being invariant to shifts of the logits, the DLR loss is rescaling invariant~\cite{pmlr-v119-croce20b}.

Adversarial robustness evaluations typically consist of performing adversarial attacks to obtain an upper bound of the robustness of a model~\cite{croce2020robustbench}. One resorts to this approximation since computing it exactly is usually intractable~\cite{carlini2019evaluating,croce2020robustbench}.

Several defenses against adversarial examples have been proposed in the literature. Such defenses tend to be designed to be robust to one specific threat model~\cite{carlini2019evaluating}. Adversarial training~\cite{madry2018towards}, and variants like ensemble adversarial training~\cite{tramer2018ensemble}, have shown the most promising results when it comes to increasing the robustness of models. The main idea behind these methods is to incorporate adversarial examples in the training procedure of a model. Many of the other methods were circumvented shortly after their proposal~\cite{pmlr-v80-athalye18a,DBLP:conf/nips/TramerCBM20}.


\subsection{Related Work}
\label{ssec:related-work}

RobustBench~\cite{croce2020robustbench} aims at benchmarking adversarial robustness under different threat models, namely $L_{p}$-robustness with $p \in \left\{2, \infty \right\}$ and common image corruptions~\cite{hendrycks2018benchmarking}. In addition to a leaderboard aggregating the evaluations of several robustness-enhancing proposals, a Model Zoo containing pre-trained models from top entries of the leaderboard is also available. While the primary goal of RobustBench is to assess defenses against adversarial attacks, we focus on the general robustness of a model from an architectural point-of-view.

In~\cite{benz2021robustness}, a comparison is made between CNNs and more recent architectures, such as the Vision Transformer~\cite{dosovitskiy2021an} and the MLP-Mixer~\cite{DBLP:journals/corr/abs-2105-01601}, which have achieved promising results in computer vision tasks. This work relates to ours in that architectural differences are at the core of the analysis, but we solely focus on CNNs and try to establish a comparison between hand-crafted models and models that result from an evolutionary search.

The experiments conducted by~\cite{Devaguptapu_2021_ICCV} are closer to our approach since the authors also look at adversarial robustness from an architectural perspective and include both manually-designed architectures and architectures found by NAS approaches (such as an NSGA-Net model) in their analysis. In contrast to this work, we focus only on NeuroEvolution methods. Moreover, we consider not only $L_{\infty}$-robustness but also $L_{2}$-robustness, and we try to perform attacks that bring the accuracy of at least one of the target models to zero under each threat model.

There is also a growing body of work that uses NAS approaches, including NE, to find robust models~\cite{DBLP:conf/cvpr/GuoYX0L20,sinn2019safeml,DBLP:journals/corr/abs-1906-11667}. However, and as pointed out by~\cite{Devaguptapu_2021_ICCV}, adversarial training is often incorporated in these studies. Thus, it is difficult to assess the role of architectural aspects on the robustness exhibited by the models.
In the case of~\cite{DBLP:journals/corr/abs-1906-11667}, robustness is explicitly included in the objective function, once again making it difficult to understand if the models found by NE are inherently more robust than the ones crafted by humans.

\section{Experimental Setup}
\label{sec:experimental-setup}

In this section, we detail the methodology followed to evaluate the adversarial robustness of the models. We conduct all the experiments on the CIFAR-10 dataset~\cite{krizhevsky2009learning}, which consists of $32 \times 32$ RGB images divided into 10 classes. The training set has 50000 images while the test set has 10000, with an equal number of examples from each class.
The original pixel values are in $\left[0,255\right]$, but we always operate on pixels modeled as real numbers by applying a pre-processing step to normalize the values to the interval $\left[0,1\right]$.

\subsection{Target Models}
We evaluate models designed by two NE approaches: DENSER~\cite{assuncao2018denser} and NSGA-Net~\cite{10.1145/3321707.3321729}. This choice was mainly based on two criteria: firstly, models had to be directly trained on the CIFAR-10 dataset, and secondly, pre-trained models (i.e., network weights) had to be publicly available, so as to introduce as little bias as possible from our end. In fact, all the attacked models are pre-trained and publicly available. We use these models directly, without re-training them, but describe some relevant differences in the training procedures. 

The WRN-28-10 architecture~\cite{Zagoruyko2016WRN}, a manually-designed wide residual network, is used as the baseline model in our experiments for a variety of reasons: its performance on CIFAR-10 is similar to that of the models from the two NE approaches, the work which proposes NSGA-Net~\cite{10.1145/3321707.3321729} also uses it as a baseline, and some of the defenses from the RobustBench leaderboard~\cite{croce2020robustbench} are based on this architecture.
In particular, we use the pre-trained model from the Model Zoo of RobustBench~\footnote{https://github.com/RobustBench/robustbench}, which was trained with the 50000 training images of CIFAR-10, without any data augmentation.
Besides converting the pixel values to $\left[0,1\right]$ as previously described, no further pre-processing is applied to the data.

In what concerns DENSER~\cite{assuncao2018denser}, we select the network that achieved the best accuracy on the CIFAR-10 test set over 10 evolutionary runs. The models resulting from 5 independent training runs of this network are publicly available~\footnote{https://github.com/fillassuncao/denser-models}, but, again, we solely attack the one with the highest test accuracy.
In these training runs, the original work used the complete training set of CIFAR-10 and applied a data augmentation method which includes padding, horizontal flipping, and random crops (similar to what is done in~\cite{10.1145/3071178.3071229}).
In addition to converting the pixel values to real numbers, the data is expected to be centered around zero before being fed to the first layer of the network. Following~\cite{assuncao2018denser}, this is accomplished through the removal of the mean pixel value per location and color channel (calculated on the entire training set).

As far as the NSGA-Net approach~\cite{10.1145/3321707.3321729} is concerned, we conduct experiments with a pre-trained model from the macro search space described in Section~\ref{ssec:neuroevolution} (NSGA-M), as well as three variants of the architecture obtained by using the cells found by NSGA-Net on the NASNet micro search space~\cite{8579005} (NSGA-mA, NSGA-mB, and NSGA-mC with an increasingly higher number of model parameters, as shown in Table~\ref{tab:parameters-and-clean-acc}).
In the original work, cutout~\cite{devries2017cutout} was used to train these models, together with a data augmentation strategy similar to the one adopted by DENSER, which includes padding, random crops, and horizontal flipping.
For the three models from the micro search space, the scheduled path dropout technique~\cite{8579005} was also adopted, together with an auxiliary head classifier, whose loss is aggregated with the loss from the main network~\cite{10.1145/3321707.3321729}.
After converting the pixel values to real numbers, the data is expected to be normalized using pre-calculated means and standard deviations for each color channel.
Further details about these architectures and training procedure can be found in the original paper~\cite{10.1145/3321707.3321729} and the source code repository~\footnote{https://github.com/ianwhale/nsga-net}.


An overview of the size of the models used in our experiments, as given by the number of trainable parameters, is presented in Table~\ref{tab:parameters-and-clean-acc}.

\begin{table}[ht]
    \renewcommand{\arraystretch}{1.2}
    \vspace{-8pt}
    \centering
    \caption{Overview of the models in terms of number of parameters and accuracy on the clean examples of the CIFAR-10 test set.}
    \label{tab:parameters-and-clean-acc}
    \begin{tabular}{c|c|c}
    \textbf{Model}   & \textbf{Number of Parameters} & \textbf{Clean Accuracy} \\
    \hline
    WRN-28-10        & 36.48M                        &  94.78\% \\
    DENSER           & 10.81M                        &  93.70\% \\
    NSGA-M           & 3.37M                         &  96.27\% \\
    NSGA-mA          & 1.97M                         &  97.57\% \\
    NSGA-mB          & 2.20M                         &  97.78\% \\
    NSGA-mC          & 4.05M                         &  97.98\% \\
    \end{tabular}
    \vspace{-8pt}
\end{table}

\subsection{Threat Models and Attacks}
Since all models are publicly available, we consider the scenario in which the attacker has full access to the target model, i.e., we perform white-box attacks. Furthermore, we consider untargeted attacks, where the adversarial perturbations are bounded by $\epsilon = 8 / 255$ under the $L_{\infty}$-norm or, in the case of the $L_{2}$-norm, by $\epsilon = 0.5$. The perturbation budgets $\epsilon$ were chosen based on threat models used in previous works, namely in the RobustBench benchmark~\cite{croce2020robustbench}.

We focus on attacks that craft adversarial examples by solving a constrained optimization problem instead of attacks that aim at finding minimal adversarial perturbations. Therefore, we attack the chosen models using different configurations (number of iterations and number of random initializations) of FGSM / FGM, BIM, and PGD. For the iterative attacks (i.e., BIM and PGD), we set the step size to $\alpha = \epsilon / 4$. We use the Python implementations of the attacks by the Adversarial Robustness Toolbox (ART) library~\cite{art2018}.

\subsection{Baseline Performance}
The accuracy of the models on the clean examples of the test set is shown in Table~\ref{tab:parameters-and-clean-acc}. For a fair comparison, we only generate adversarial examples on samples that initially receive a correct classification by the model under evaluation.
Nevertheless, when reporting the models' accuracy of adversarially generated samples, we consider the complete test set of CIFAR-10. It is important to mention that an untargeted attack is considered to be successful if the model produces a misclassification, regardless of the predicted class. For this reason, no perturbation needs to be added to a sample that is already incorrectly classified.

While the DENSER models are implemented in Keras / TensorFlow 2, the baseline and the NSGA-Net models are in PyTorch. We reiterate that all models were trained using a standard procedure and no defensive method was applied.

\section{Results and Discussion}
\label{sec:results-discussion}

The accuracy of the models on the adversarial examples generated under the threat model that considers $L_{\infty}$-robustness is shown in Table~\ref{tab:results-common}. We present the results for the FGSM attack, for FGSM with 10 random initializations (FGSM-10), and for the BIM attack with 10 and 50 iterations (BIM-10 and BIM-50, respectively). In this case, the attacks operate in $\left[0,1\right]$ and any additional model-specific pre-processing is applied to the images after the adversarial perturbations are added.

A brief perusal of the results reveals that the DENSER model is the most susceptible to $L_{\infty}$ attacks. Even in the case of single-step attacks like FGSM, the accuracy falls below 10\% when random restarts are incorporated. On the other hand, the models that result from the application of NSGA-Net to the NASNet search space are the most resistant to single-step attacks. In fact, they achieve higher accuracy on the adversarially perturbed images than the baseline model. Nevertheless, given enough iterations, the accuracy of all models drops to zero under this threat model. This suggests that these NE approaches do not seem to find $L_{\infty}$-robust models, at least if that objective is not explicitly included in the evolutionary process.

Table~\ref{tab:results-common} also shows the adversarial accuracy of the models under the threat model that considers $L_{2}$-bounded perturbations. We present the results for the FGM attack, for the BIM attack with 10, 50, and 100 steps (BIM-10, BIM-50, and BIM-100, respectively), as well as for the PGD attack with 10 random restarts and 50 iterations (PGD-50-10). We again report the results when the attacks operate in $\left[0,1\right]$ and any additional model-specific pre-processing is applied after the adversarial perturbations are added to the images.

\begin{table*}[ht]
    \renewcommand{\arraystretch}{1.1}
    \centering
    \caption{Accuracy on the CIFAR-10 test set when the attacks operate in $\left[0,1\right]$. The highest reported accuracy under each attack is in bold.}
    \label{tab:results-common}
    \vspace{-3pt}
        \begin{tabular}{c|c|cccccc}
            \multicolumn{2}{c|}{}                                                    & \textbf{WRN-28-10} & \textbf{DENSER} & \textbf{NSGA-M} & \textbf{NSGA-mA} & \textbf{NSGA-mB} & \textbf{NSGA-mC}\\
            \hline
            \multirowcell{4}{$L_{\infty}$ \\ $\epsilon = 8 / 255$} & FGSM            & 28.85\%             & 16.37\%          & 35.08\%              & 52.09\%               & 51.86\%               & \textbf{55.06\%}    \\ 
                                                                   & FGSM-10         & 11.03\%             &  6.19\%          &  9.28\%              & 25.02\%               & 22.49\%               & \textbf{26.92\%}    \\
                                                                   & BIM-10          &  0.02\%             &  0.00\%          &  0.00\%              & \textbf{0.16\%}       &  0.00\%               &  0.02\%             \\
                                                                   & BIM-50          & \textbf{0.00\%}     & \textbf{0.00\%}  & \textbf{0.00\%}      & \textbf{0.00\%}       & \textbf{0.00\%}       & \textbf{0.00\%}    \\
            \hline
            \multirowcell{5}{$L_{2}$ \\ $\epsilon = 0.5$}          & FGM             & 47.61\%             & 44.76\%          & 48.51\%              & 61.34\%               & 60.61\%               & \textbf{64.06\%}    \\ 
                                                                   & BIM-10          &  2.01\%             & \textbf{30.76\%} &  0.23\%              &  3.04\%               &  0.73\%               &  2.57\%             \\
                                                                   & BIM-50          &  0.16\%             & \textbf{24.13\%} &  0.00\%              &  0.26\%               &  0.01\%               &  0.35\%             \\
                                                                   & BIM-100         &  0.09\%             & \textbf{21.76\%} &  0.00\%              &  0.12\%               &  0.00\%               &  0.23\%             \\
                                                                   & PGD-50-10       &  0.08\%             & \textbf{18.10\%} &  0.00\%              &  0.11\%               &  0.00\%               &  0.21\%             \\
        \end{tabular}
    \vspace{-10pt}
\end{table*}

\begin{table*}[ht]
    \renewcommand{\arraystretch}{1.1}
    \centering
    \caption{Accuracy on the CIFAR-10 test set when all model-specific pre-processing is applied to the original inputs before performing the attacks. The highest reported accuracy for each attack is in bold.}\label{tab:results-model-specific}
        \begin{tabular}{c|c|ccccc}
            \multicolumn{2}{c|}{}                                                    & \textbf{DENSER} & \textbf{NSGA-M} & \textbf{NSGA-mA} & \textbf{NSGA-mB} & \textbf{NSGA-mC} \\
            \hline
            \multirowcell{4}{$L_{\infty}$ \\ $\epsilon = 8 / 255$} & FGSM            & 16.37\%          & 46.65\%              & 60.61\%               & 61.57\%               & \textbf{63.64\%}      \\ 
                                                                   & FGSM-10         &  6.16\%          & 40.82\%              & 58.41\%               & 58.88\%               & \textbf{60.97\%}      \\
                                                                   & BIM-10          &  0.00\%          &  2.70\%              & 12.22\%               &  8.36\%               & \textbf{12.70\%}      \\
                                                                   & BIM-50          &  0.00\%          &  0.81\%              &  6.87\%               &  3.86\%               & \textbf{6.96\%}      \\
            \hline  
            \multirowcell{5}{$L_{2}$ \\ $\epsilon = 0.5$}          & FGM             & 44.75\%          & 67.96\%              & 78.11\%               & 77.70\%               & \textbf{80.12\%}      \\ 
                                                                   & BIM-10          & 30.77\%          & 25.98\%              & 48.60\%               & 44.80\%               & \textbf{49.61\%}      \\
                                                                   & BIM-50          & 24.13\%          & 17.57\%              & 41.63\%               & 37.43\%               & \textbf{42.04\%}      \\
                                                                   & BIM-100         & 21.76\%          & 16.86\%              & 40.72\%               & 36.46\%               & \textbf{41.09\%}      \\
                                                                   & PGD-50-10       & 18.10\%          & 15.99\%              & \textbf{40.10\%}      & 35.38\%               & 40.05\%               \\
        \end{tabular}
    \vspace{-10pt}
\end{table*}

\begin{table*}[ht]
    \renewcommand{\arraystretch}{1.1}
    \centering
    \caption{Accuracy on the CIFAR-10 test set when the attacks operate in $\left[0,1\right]$, but the generated images are post-processed. The highest reported accuracy for each attack is in bold.}
    \label{tab:results-post}
        \begin{tabular}{c|c|cccccc}
            \multicolumn{2}{c|}{}                                                    & \textbf{WRN-28-10} & \textbf{DENSER} & \textbf{NSGA-M} & \textbf{NSGA-mA} & \textbf{NSGA-mB} & \textbf{NSGA-mC}\\
            \hline
            \multirowcell{4}{$L_{\infty}$ \\ $\epsilon = 8 / 255$} & FGSM            & 28.85\%             & 16.38\%          & 35.08\%              & 52.09\%               & 51.86\%               & \textbf{55.06\%}    \\ 
                                                                   & FGSM-10         & 11.15\%             &  6.28\%          &  9.45\%              & 25.19\%               & 22.70\%               & \textbf{27.13\%}    \\
                                                                   & BIM-10          &  0.02\%             &  0.00\%          &  0.00\%              & \textbf{0.16\%}       & 0.00\%                & 0.02\%              \\
                                                                   & BIM-50          & \textbf{0.00\%}     & \textbf{0.00\%}  & \textbf{0.00\%}      & \textbf{0.00\%}       & \textbf{0.00\%}       & \textbf{0.00\%}     \\
            \hline
            \multirowcell{5}{$L_{2}$ \\ $\epsilon = 0.5$}          & FGM             & 47.69\%             & 44.86\%          & 48.62\%              & 61.46\%               & 60.82\%               & \textbf{64.24\%}    \\ 
                                                                   & BIM-10          &  2.02\%             & \textbf{30.77\%} &  0.23\%              &  3.10\%               &  0.75\%               &  2.67\%             \\
                                                                   & BIM-50          &  0.16\%             & \textbf{24.13\%} &  0.00\%              &  0.26\%               &  0.01\%               &  0.38\%             \\
                                                                   & BIM-100         &  0.09\%             & \textbf{21.76\%} &  0.00\%              &  0.12\%               &  0.00\%               &  0.24\%             \\
                                                                   & PGD-50-10       &  0.08\%             & \textbf{18.10\%} &  0.00\%              &  0.14\%               &  0.01\%               &  0.23\%             \\
        \end{tabular}
    \vspace{-10pt}
\end{table*}

The strongest $L_{2}$ attacks in our analysis (BIM-100 and PGD-50-10) bring the accuracy of the models to below 1\% and, in the case of NSGA-M and NSGA-mB, to zero. Surprisingly, and contrary to what was observed with the $L_{\infty}$ attacks, this does not hold true for the DENSER model whose accuracy just drops to around 20\% under this threat model.
However, the DENSER model remains the most susceptible under the single-step FGM attack, while the NSGA-Net models from the search space of NASNet remain the most robust.

Moreover, a comparison between the three NSGA-Net models from the micro search space reveals that NSGA-mB is the least robust of the three, even though it is more complex (i.e., it has a higher number of parameters) than NSGA-mA.
This is observed under both distance metrics, but the susceptibility of NSGA-mB is particularly higher than that of NSGA-mA and NSGA-mC under the BIM-10 attack with $L_{2}$ perturbations.
Unlike NSGA-mA and NSGA-mC, NSGA-mB does not use Squeeze-and-Excitation blocks. Thus, it seems that Squeeze-and-Excitation may help improve the robustness of an ANN, an hypothesis worth investigating in future work.
The discrepancies in the relative robustness of the models between the two distance metrics demands further analysis. Namely, it would be of interest to understand what aspect of the DENSER model makes it more $L_{2}$-robust and why it does not seem to help the model against $L_{\infty}$ attacks.

\subsection{Impact of Data Pre-Processing}
In Section~\ref{sec:experimental-setup}, we described the different pre-processing steps applied to the images before they reach the first layer of each model. Contrary to WRN-28-10 and DENSER, the pre-processing for the NSGA-Net models changes the scale of the data (i.e., the difference between the maximum and the minimum values of a feature is larger than 1). Therefore, the NSGA-Net models perceive the adversarial perturbations as approximately 4 times larger than the perturbation budget of the threat model.
To show this effect, we craft adversarial examples after all the pre-processing steps have been applied to the data, instead of operating in the $\left[0,1\right]$ range. By doing so, the perturbation budget refers to the distance in the space in which the first layer of a network expects the data to be.

The results of this experiment are shown in Table~\ref{tab:results-model-specific} for both the $L_{\infty}$ and the $L_{2}$ attacks. The baseline model is excluded from this analysis since it only requires the data to be in $\left[0,1\right]$, without centering or standardizing it. We can see that, as expected, there is no significant difference between these results and those from Table~\ref{tab:results-common} regarding the DENSER model. On the other hand, the robustness of the NSGA-Net models appears to be much higher, especially in the case of $L_{2}$-bounded perturbations. This shows that the choice of data pre-processing should not be neglected when designing networks under scenarios where adversarial attacks may be of concern. In the particular case of NE approaches, one might even consider including this design choice in the evolutionary process. Additionally, works that focus on robustness evaluations should clearly specify the conditions under which the models were trained, including any model-specific pre-processing step.

We also evaluate the impact of converting the pixel values back to integers, such that each value is between 0 and 255. We just consider the case in which the attacks operate in the range $\left[0,1\right]$ and any additional pre-processing is applied after the perturbations have been added to the images. Therefore, we multiply each pixel value by 255 and round to the nearest even integer. We then re-apply all pre-processing steps required by the model and report the accuracy on the post-processed examples. Table~\ref{tab:results-post} shows the results for the attacks under both distance metrics.

For the $L_{\infty}$ attacks, differences are mainly detected when random restarts are incorporated (FGSM-10). The attack success rate slightly deteriorates, but the differences are of less than 0.25\% and seem negligible. As far as the $L_{2}$-bounded attacks are concerned, the largest differences occur with FGM but also seem negligible (always of less than 0.25\%). The success of the attacks is mostly affected by this post-processing procedure when their target is an NSGA-Net model from the NASNet search space.

\begin{figure}[!t]
    \centering
    \subfloat[BIM-50, $L_{\infty}$-constrained]{
        \includegraphics[width=0.75\columnwidth]{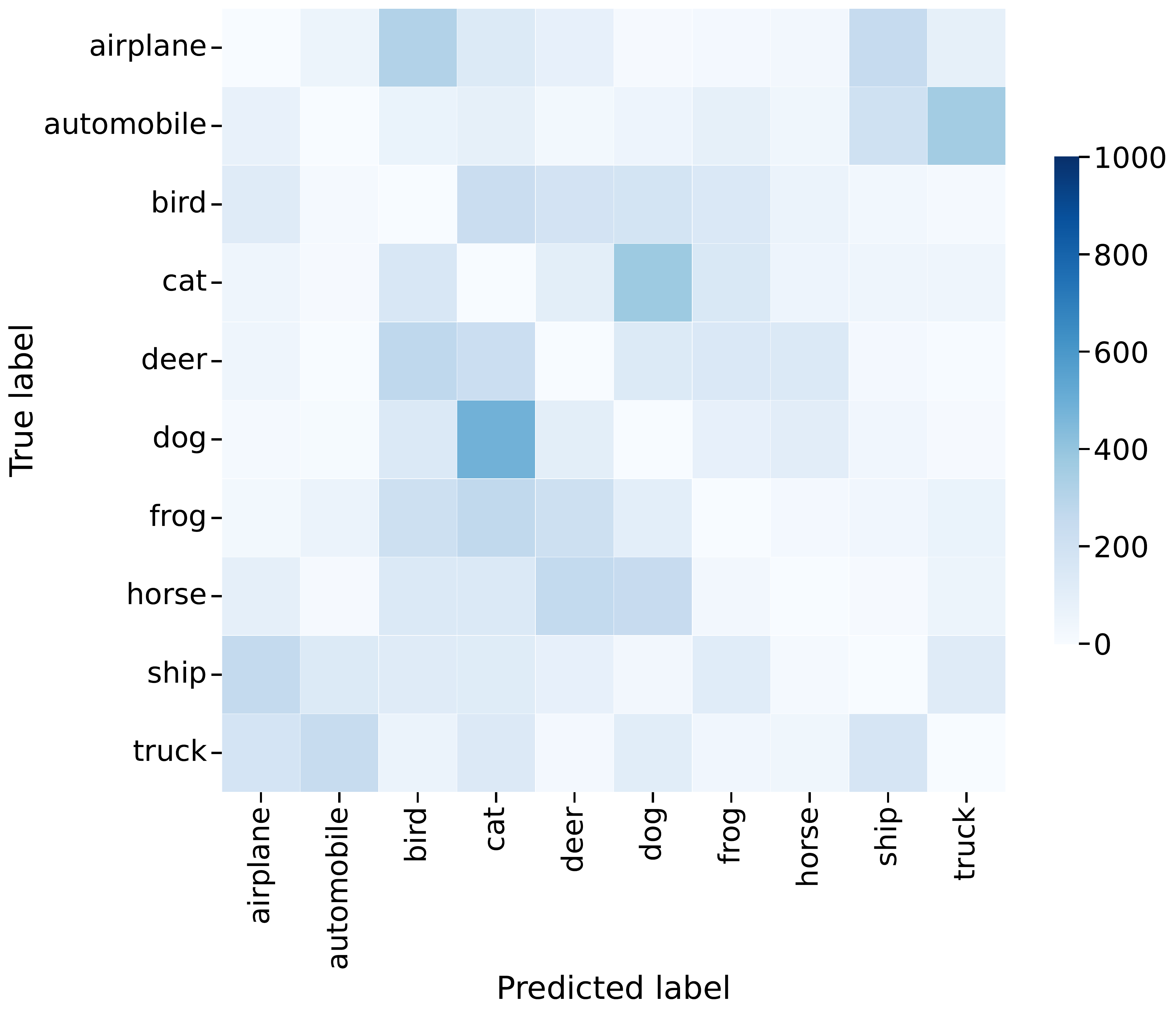}
        \label{fig:cm-linf-bim50-wrn}
    }\\
    \subfloat[BIM-100, $L_{2}$-constrained]{
        \includegraphics[width=0.75\columnwidth]{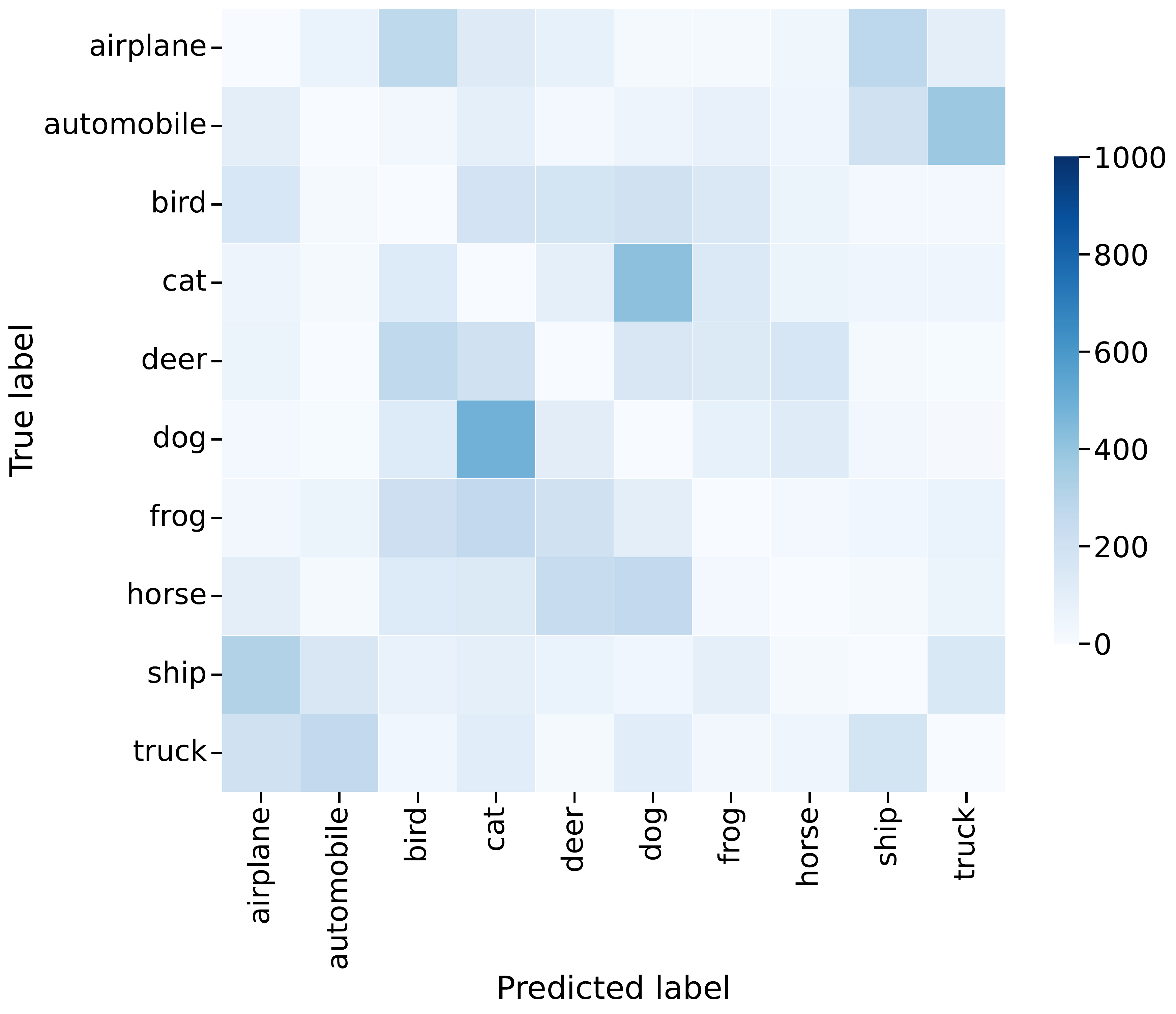}
        \label{fig:cm-l2-bim100-wrn}
    }
    \caption{Confusion matrices for the WRN-28-10 model under two attacks.}
    \label{fig:cm-wrn}
    \vspace{-10pt}
\end{figure}

\subsection{Analyzing Misclassifications through Confusion Matrices}
To complement our analysis, we looked into the confusion matrices of each model under different attacks. Even when an attack brings the accuracy of all models to zero (e.g., the $L_{\infty}$-constrained BIM-50 attack), different patterns in their misclassifications can be observed.

As shown in Fig.~\ref{fig:cm-linf-bim50-wrn}, under an $L_{\infty}$-bounded BIM-50 attack, WRN-28-10 produces misclassifications for each class that are spread out across the remaining classes. Moreover, it seems to favor mainly two classes, bird and cat, with most examples being misclassified as such.
Under the $L_{2}$-constrained BIM-100 attack (Fig.~\ref{fig:cm-l2-bim100-wrn}), the misclassifications of classes that represent a means of transportation (airplane, automobile, ship, and truck) are more clustered together.

\begin{figure}[!t]
    \centering
    \subfloat[BIM-50, $L_{\infty}$-constrained]{
        \includegraphics[width=0.75\columnwidth]{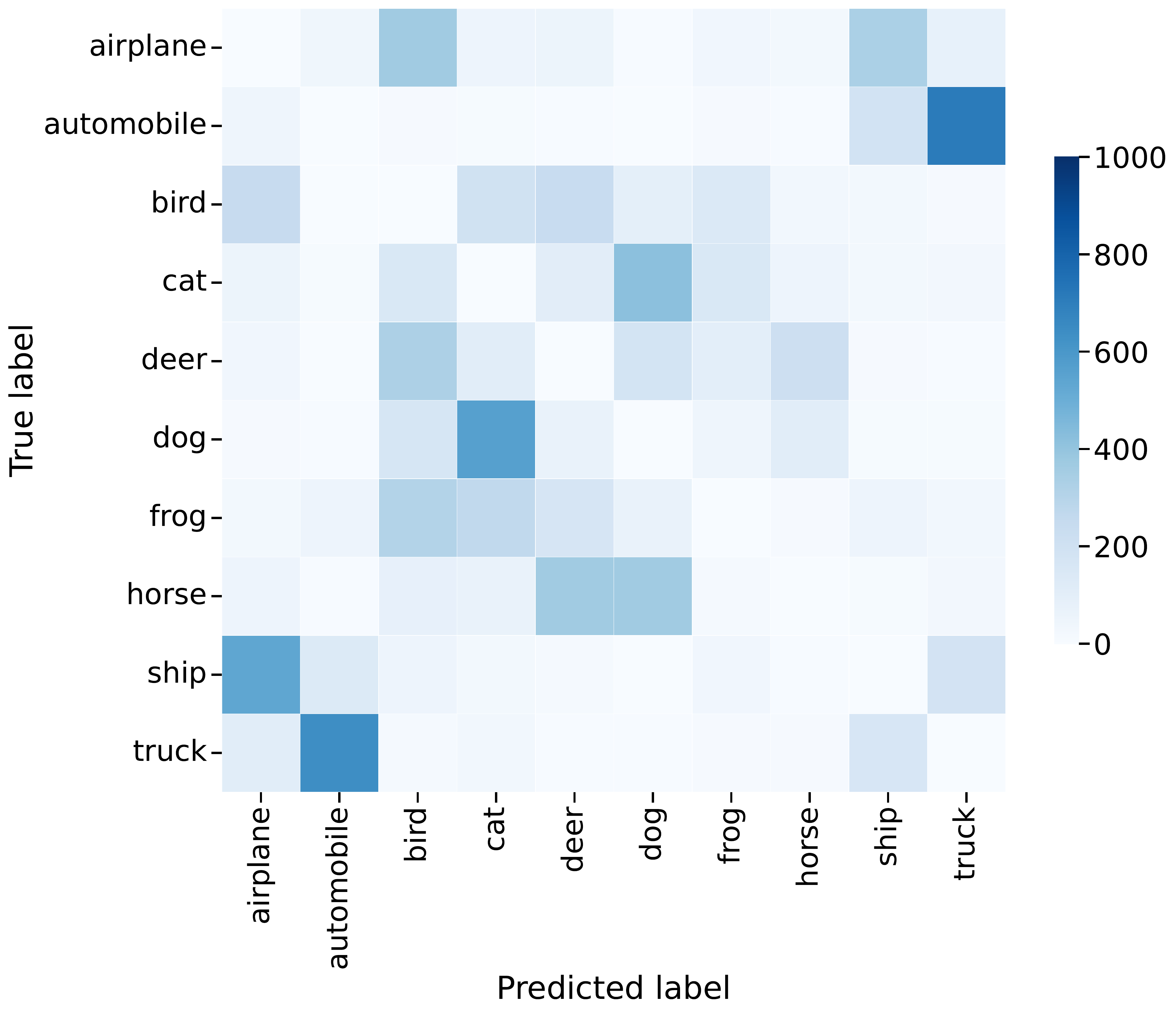}
        \label{fig:cm-linf-bim50-denser}
    }\\
    \subfloat[BIM-100, $L_{2}$-constrained]{
        \includegraphics[width=0.75\columnwidth]{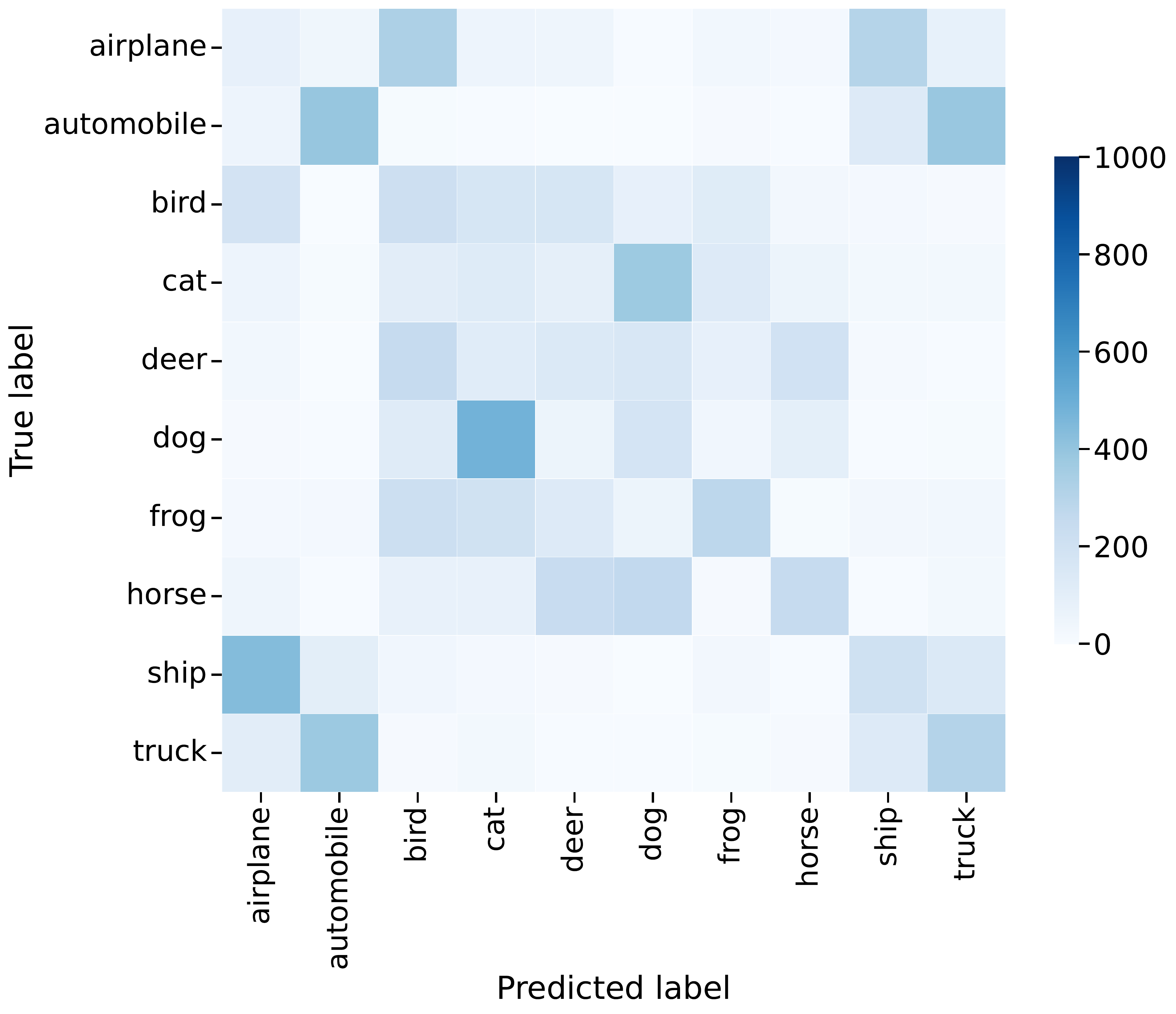}
        \label{fig:cm-l2-bim100-denser}
    }
    \caption{Confusion matrices for the DENSER model under two attacks.}
    \label{fig:cm-denser}
    \vspace{-10pt}
\end{figure}

Fig.~\ref{fig:cm-linf-bim50-denser} shows that, under the $L_{\infty}$-constrained BIM-50 attack, the predictions of the DENSER model can be clearly grouped into clusters, with most examples from one class being misclassified into a smaller subset of the other classes than with the baseline model. Images that represent an animal are mainly misclassified as another animal, while images of a means of transportation are misclassified as another vehicle.
The confusion matrix of DENSER under the BIM-100 attack constrained by the $L_{2}$-norm is shown in Fig.~\ref{fig:cm-l2-bim100-denser}. Contrary to the $L_{\infty}$ attack, the BIM-100 attack is unable to decrease the accuracy to zero, and so, some images are correctly classified. Notwithstanding, the misclassifications follow a pattern similar to that shown in Fig.~\ref{fig:cm-linf-bim50-denser}. The automobile and the truck classes are the most difficult to attack under this threat model, while it is easier to cause a misclassification of airplane instances.

\begin{figure*}[!th]
    \centering
    \subfloat[NSGA-M]{
        \includegraphics[width=0.75\columnwidth]{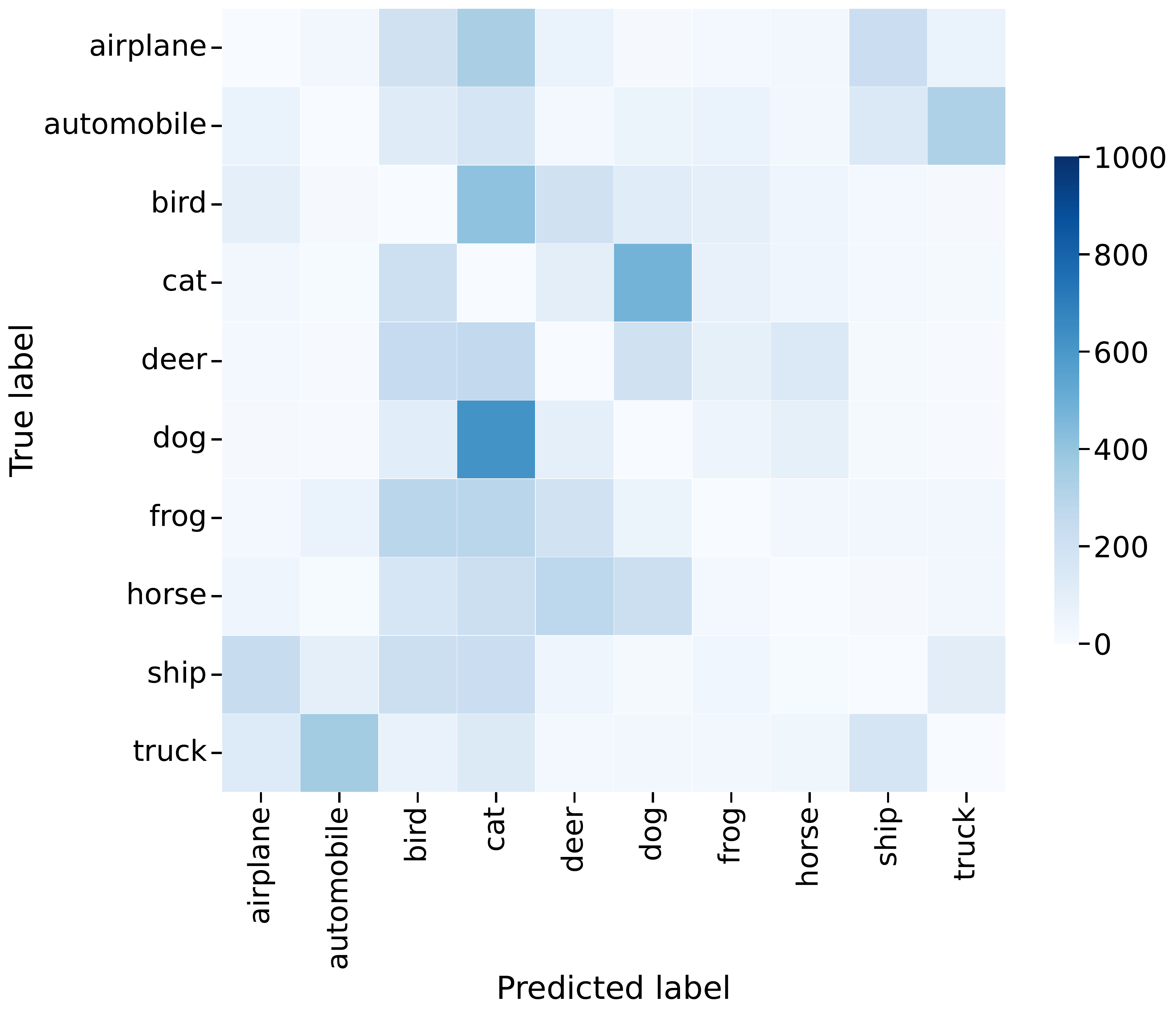}
        \label{fig:cm-linf-bim50-macro}
    }\hspace{10mm}
    \subfloat[NSGA-mA]{
        \includegraphics[width=0.75\columnwidth]{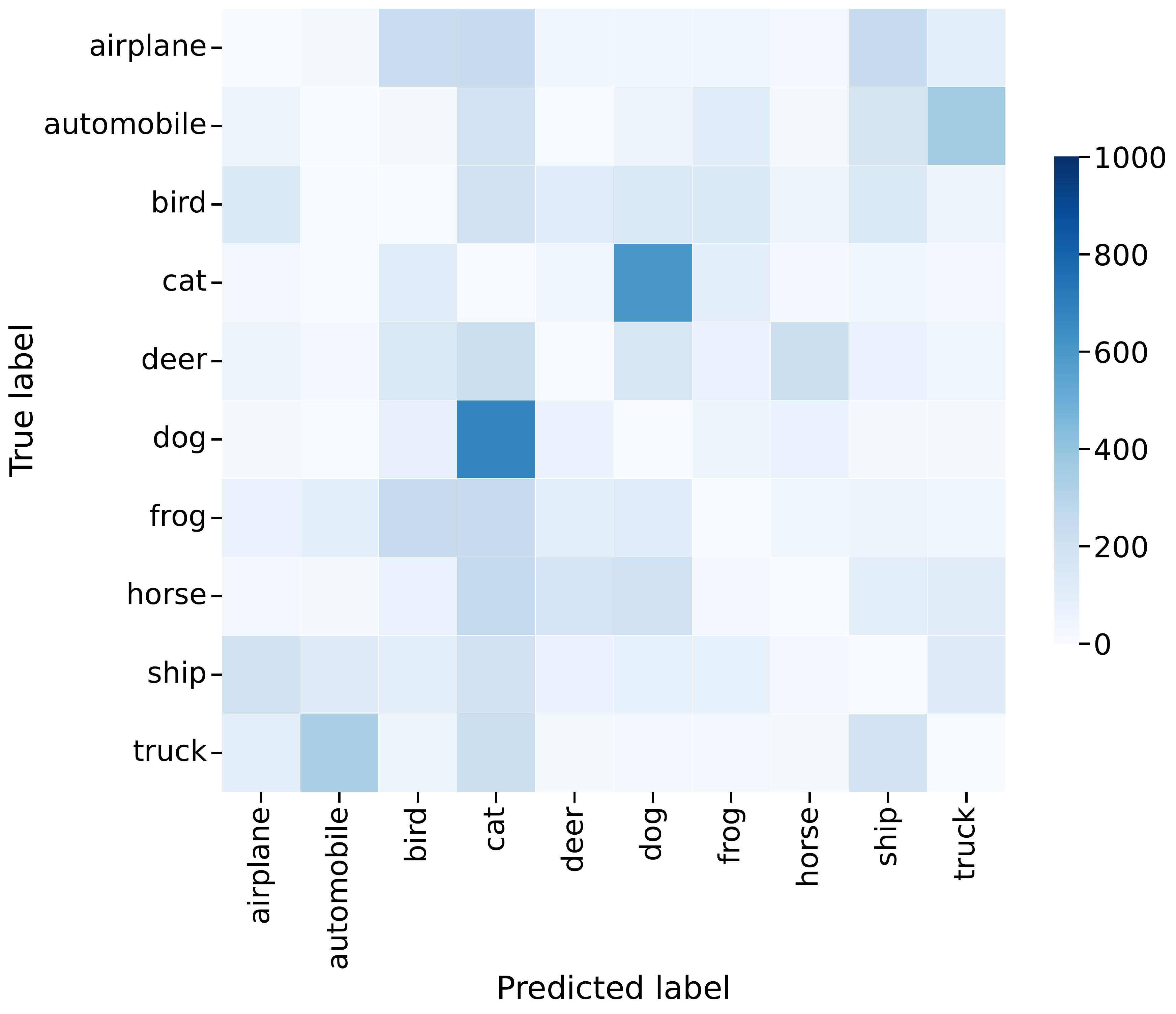}
        \label{fig:cm-linf-bim50-microA}
    }\\
    \subfloat[NSGA-mB]{
        \includegraphics[width=0.75\columnwidth]{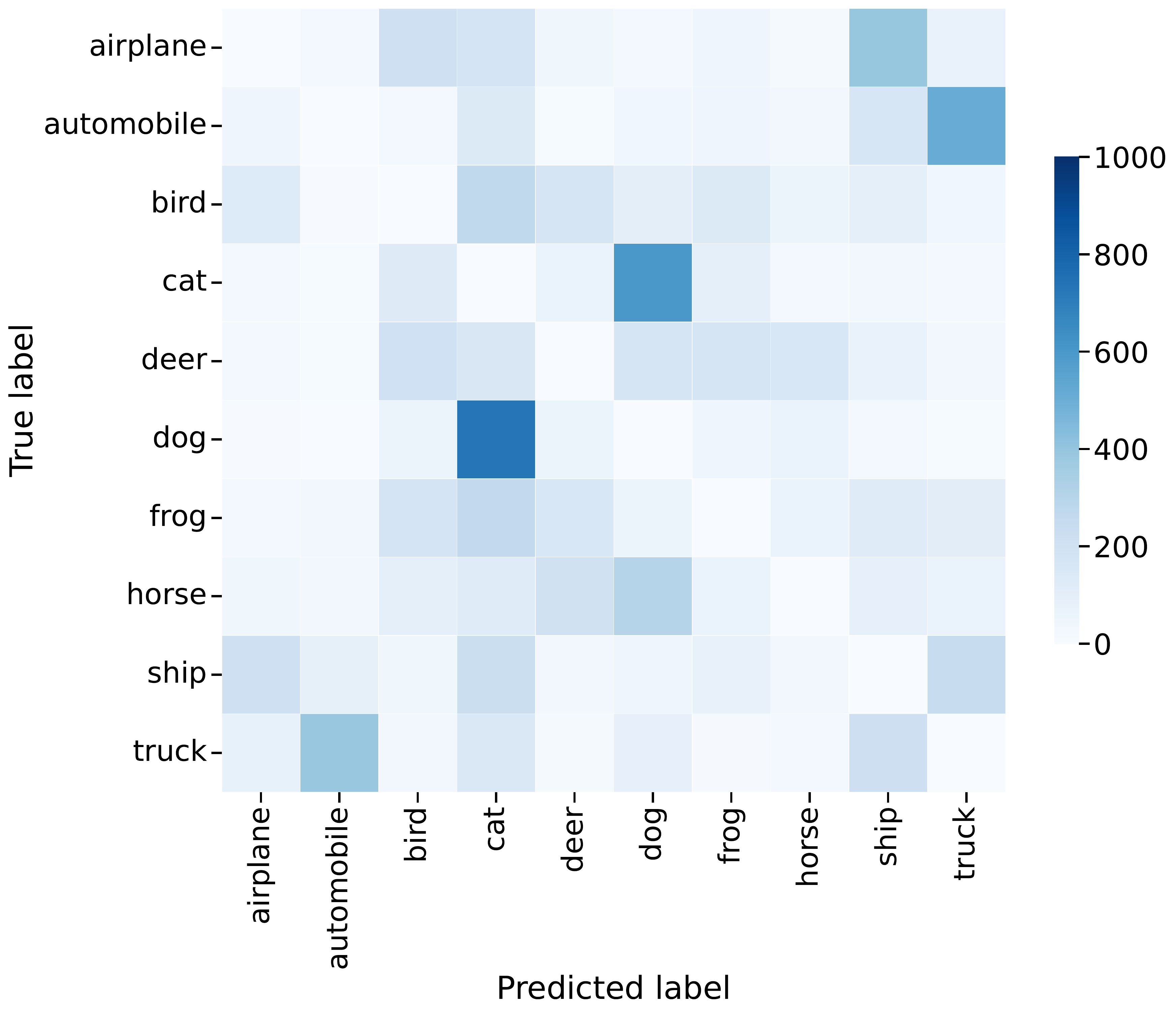}
        \label{fig:cm-linf-bim50-microB}
    }\hspace{10mm}
    \subfloat[NSGA-mC]{
        \includegraphics[width=0.75\columnwidth]{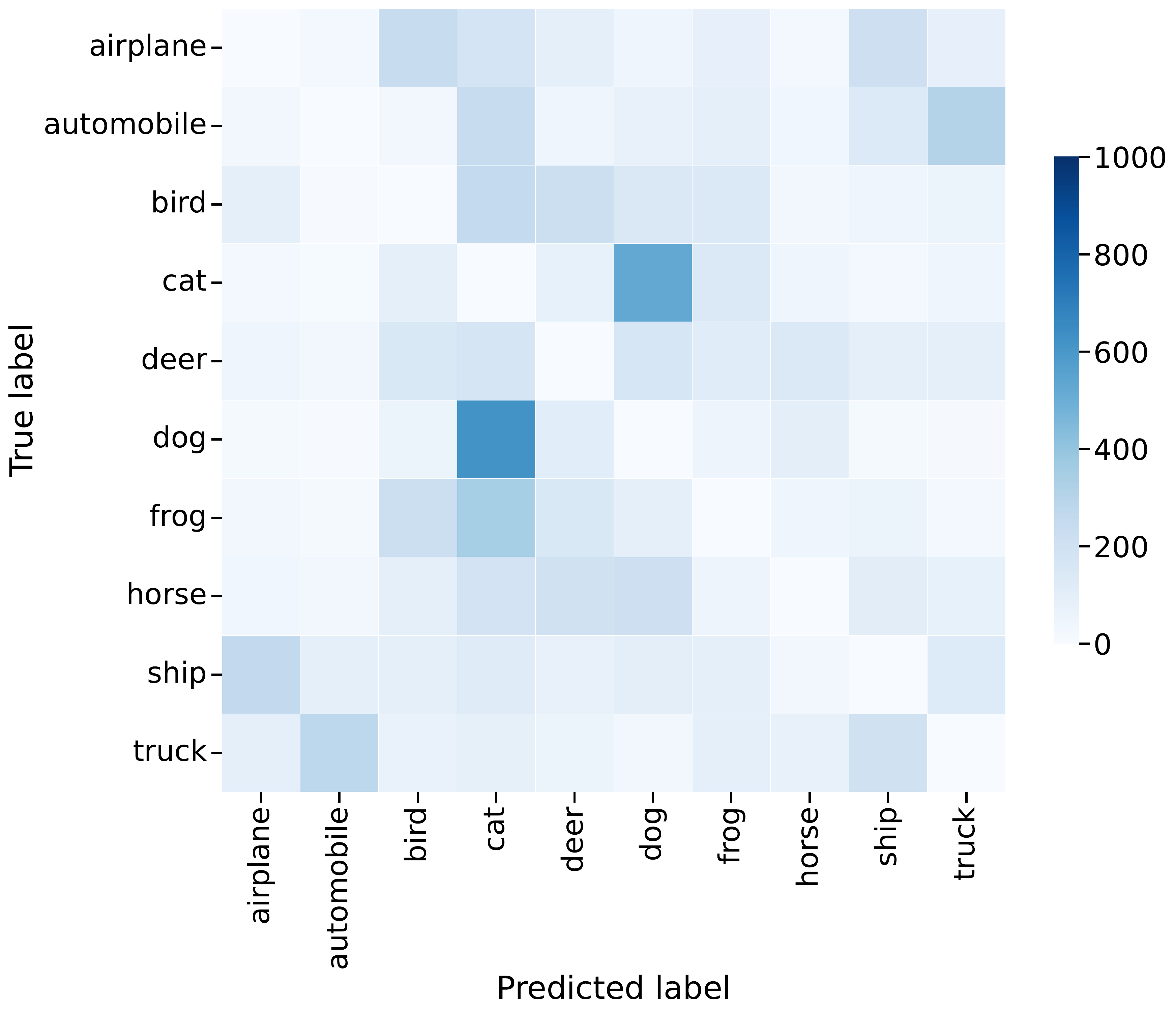}
        \label{fig:cm-linf-bim50-microC}
    }
    \caption{Confusion matrices for the NSGA-Net models under the BIM-50 attack with $L_{\infty}$-bounded perturbations. The results under BIM-100 with $L_{2}$-bounded perturbations revealed similar patterns.}
    \label{fig:cm-linf-bim50-nsga}
    \vspace{-4pt}
\end{figure*}

According to Fig.~\ref{fig:cm-linf-bim50-macro}, and similar to the baseline model, most examples are also misclassified as bird and cat with NSGA-M. However, misclassifications of a single class are less spread out between the remaining classes, especially in the case of bird, cat, ship, and truck.
The three NSGA-Net models from the NASNet search space show similar patterns in their misclassifications. The main distinguishing factor is the spread of the misclassifications of each class: NSGA-mB misclassifies the majority of the examples from one class into fewer classes than NSGA-mC (check, for instance, the ship class), and NSGA-mA is in the middle of the spectrum. Similar to NSGA-M, most misclassifications of these three models also fall into the cat class (especially with NSGA-mA). However, the second most predicted class is dog and not bird.

The confusion matrices for the BIM-100 attack with $L_{2}$-bounded perturbations exhibit similar patterns. In comparison with the BIM-50 attack with $L_{\infty}$ constraints, less examples are misclassified by the three models from the micro search space as belonging to the cat class, especially in the case of NSGA-mB. With NSGA-M and NSGA-mB, some changes are also observed with the misclassification of examples that originally belong to the ship class.

\section{Conclusion and Future Work}
\label{sec:conclusion}

Artificial Neural Networks designed through evolution achieve competitive results with respect to predictive performance, but the study of their adversarial robustness is limited.
In this work, we assessed the $L_{\infty}$ and $L_{2}$-robustness of models found by NeuroEvolution approaches for the CIFAR-10 classification task, under white-box untargeted attacks.
No defense against adversarial examples was incorporated in the models or in their training.

Our results show that, similar to human-designed networks, the accuracy of these evolved models usually drops to zero (or close to zero).
The main exception occurs with the DENSER model, which shows some resistance to $L_{2}$ attacks, with the accuracy dropping to 18.10\% under a PGD attack. 
We identified distinct patterns in the misclassifications produced by the models: in some cases, the adversarial examples from one class are misclassified into a small subset of classes, while with other models the misclassifications are much more spread out between classes.
Furthermore, the choice of data pre-processing techniques must not be neglected when automatically designing CNNs. We have shown that certain procedures can exacerbate the adversarial perturbations before they reach the first layer of the network, this way potentially jeopardizing robustness. However, extending current NE approaches so as to include this design choice in their search is not always straightforward (i.e., NSGA-Net).

We plan on studying the $L_{2}$-robustness of the DENSER model so as to understand if it can be attributed to some architectural feature. That invaluable knowledge could be leveraged to build robust models.
We tried to be as faithful as possible to the original works under analysis, which came with some disadvantages. Namely, each model was trained under slightly different configurations, including distinct data pre-processing approaches. It would be interesting to re-train all the models under the exact same conditions so as to make sure that the observed differences truly arise from architectural aspects.
Although previous work suggests that higher network capacity allows for robustness improvements, analyzing the relationship between the adversarial robustness and the computational complexity of a model still needs further investigation.

Since we solely analyzed pre-trained models for CIFAR-10, a clear extension to our work would be to consider other, more complex, datasets. Future work also comprises assessing the adversarial robustness of models found by NE approaches under additional threat models, such as transfer and universal attacks.

\section*{Acknowledgments}

This work is partially funded by national funds through the FCT - Foundation for Science and Technology, I.P., within the scope of the project CISUC - \texttt{UID/CEC/00326/2020} and by European Social Fund, through the Regional Operational Program Centro 2020.
It is also partially supported by the project METRICS (\texttt{POCI-01-0145-FEDER-032504}), co-funded by FCT and by the 
\textit{Fundo Europeu de Desenvolvimento Regional} (FEDER) through 
\textit{Portugal 2020 - Programa Operacional Competitividade e Internacionalização (POCI)}.
The first author is partially funded by FCT under the individual grant \texttt{UI/BD/151047/2021}.

\linespread{0.97}
\bibliographystyle{IEEEtran}
\bibliography{IEEEabrv,references}

\end{document}